\def\paperlanguage{}
\pdfoutput=1 

\documentclass{tADR2e}

\usepackage{bm}
\usepackage{cite}
\include{preamble}

\newcommand{\ctext}[1]{\raise0.2ex\hbox{\textcircled{\scriptsize{#1}}}}

\title{\LARGE \textbf
  {
    \switchlanguage%
    {%
      Reflex-Based Open-Vocabulary Navigation\\ without Prior Knowledge Using Omnidirectional Camera and Multiple Vision-Language Models
    }%
    {%
      全天球カメラと事前学習済み視覚-言語モデルによる事前知識を必要としない反射型Open Vocabulary Navigation
    }%
  }
}

\author{Kento Kawaharazuka$^{a}$$^{\ast}$, Yoshiki Obinata$^{a}$, Naoaki Kanazawa$^{a}$, Naoto Tsukamoto$^{a}$,\\Kei Okada$^{a}$, and Masayuki Inaba$^{a}$
    \thanks{$^\ast$Corresponding author. Email: kawaharazuka@jsk.imi.i.u-tokyo.ac.jp \vspace{6pt}}\\\vspace{6pt}  $^{a}${\em{The Department of Mechano-Informatics, Graduate School of Information Science and Technology, The University of Tokyo, 7-3-1 Hongo, Bunkyo-ku, Tokyo, Japan.}}
  }
\begin{document}

\jvol{00} \jnum{00} \jyear{2024} \jmonth{August}

\maketitle

\begin{abstract}
  \switchlanguage%
  {%
    Various robot navigation methods have been developed, but they are mainly based on Simultaneous Localization and Mapping (SLAM), reinforcement learning, etc., which require prior map construction or learning.
    In this study, we consider the simplest method that does not require any map construction or learning, and execute open-vocabulary navigation of robots without any prior knowledge to do this.
    We applied an omnidirectional camera and pre-trained vision-language models to the robot.
    The omnidirectional camera provides a uniform view of the surroundings, thus eliminating the need for complicated exploratory behaviors including trajectory generation.
    By applying multiple pre-trained vision-language models to this omnidirectional image and incorporating reflective behaviors, we show that navigation becomes simple and does not require any prior setup.
    Interesting properties and limitations of our method are discussed based on experiments with the mobile robot Fetch.
  }%
  {%
    これまで, 多様なロボットのナビゲーション方法が開発されてきたが, それらは主にSLAMや強化学習等に基づくものであり, 事前マップ構築や学習を必要とする.
    そこで本研究では, 一切のマップ構築や学習等は行わず, 事前知識無しにOpen Vocabularyでナビゲーションを行う, 最もシンプルな方法を考える.
    そこで, 全天球画像と事前学習済み視覚-言語モデルを応用した.
    全天球画像は周囲を一様に見ることができるため, 軌道生成含めた煩雑な探索行動が必要なくなる.
    この全天球画像に対して複数の事前学習済み視覚-言語モデルを適用し, 反射型の行動を組み込むことで, 事前のセットアップを一切必要とせず, シンプルにナビゲーションが可能であることを示す.
    本手法の興味深い特性と限界について, モバイルロボットFetchによる実機実験から議論する.
  }%
\end{abstract}


\section{Introduction}\label{sec:introduction}
\begin{figure}[t]
  \centering
  \includegraphics[width=0.95\columnwidth]{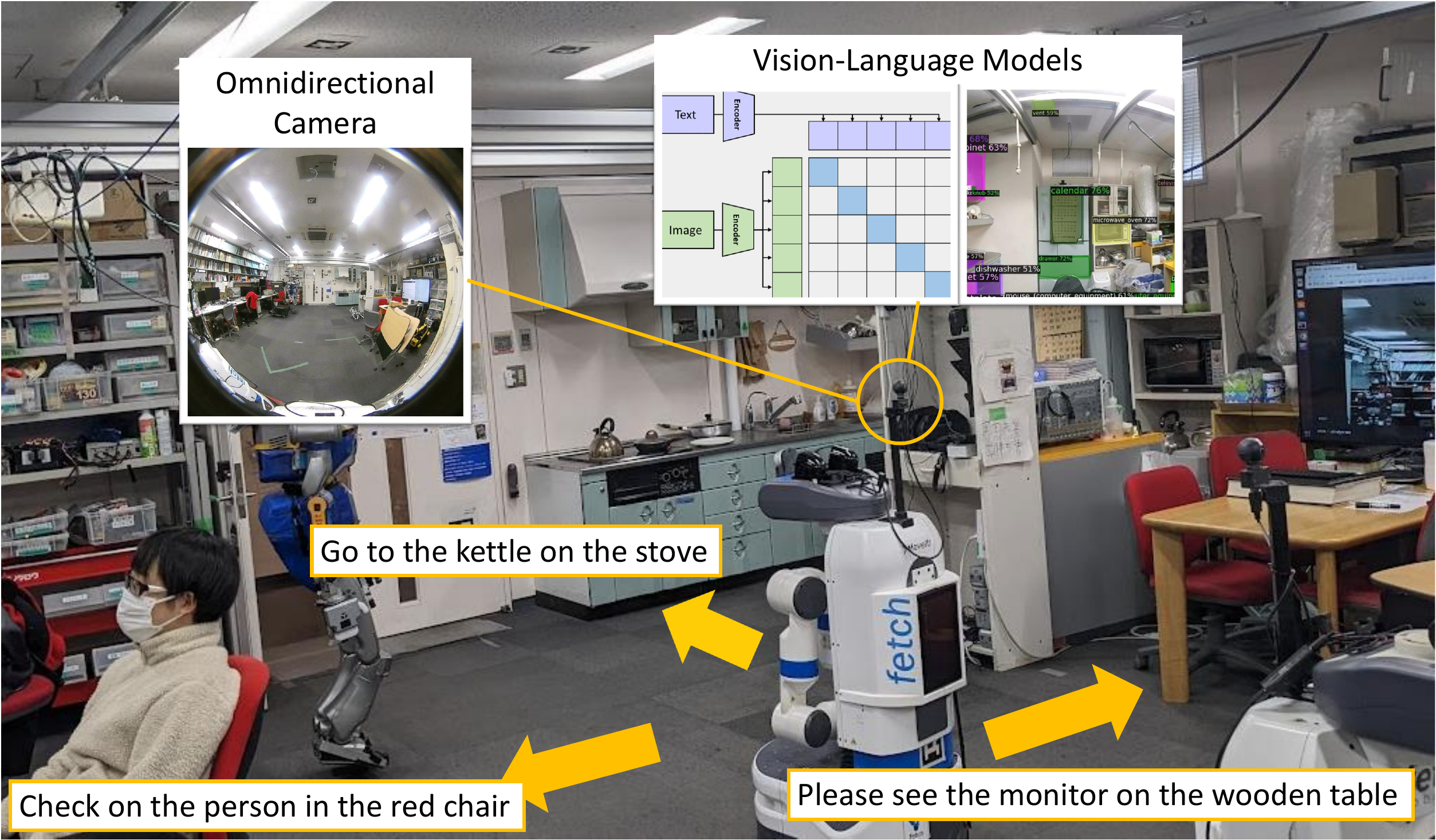}
  \caption{The concept of this study: simple reflex-based open-vocabulary navigation is enabled by splitting the expanded omnidirectional image and applying multiple pre-trained large-scale vision-language models.}
  \label{figure:concept}
\end{figure}

\switchlanguage%
{%
  Various navigation methods for robots have been developed so far.
  These methods often require prior learning and data collection, primarily through techniques such as Simultaneous Localization and Mapping (SLAM) and reinforcement learning.
  On the other hand, due to the high adaptability and responsiveness, behavior-based robot navigation methods have been developed for a long time.
  In this study, we consider performing open-vocabulary reflex-based navigation by combining these behavior-based navigation approaches with the advancements in current pre-trained vision-language models.
  We aim to achieve navigation in its simplest form without any prior learning, SLAM, or similar techniques, as illustrated in \figref{figure:concept}.
  For this purpose, we leverage omnidirectional cameras.
  By segmenting omnidirectional images and applying pre-trained vision-language models, we enable a reflex-based control that moves in the most appropriate direction according to instructions, eliminating the need for complex exploration actions including trajectory generation.
  Furthermore, we demonstrate that using multiple pre-trained vision-language models allows for more suitable navigation.
  Unlike previous research where maps or policies were generated in advance, this study intentionally avoids these steps and focuses on discussing how to achieve open-vocabulary navigation in the simplest manner possible.
  While each process itself is not new, combining them enables novel open-vocabulary reflex-based navigation.

  This study is organized as follows.
  In \secref{sec:related}, we discuss related navigation research based on map generation, reinforcement learning, and omnidirectional cameras.
  In \secref{sec:proposed}, we describe the expansion of the omnidirectional image, the application of the multiple vision-language models, and the reflex-based control based on linguistic instructions.
  In \secref{sec:experiment}, we describe quantitative evaluation experiments and more practical advanced experiments.
  In \secref{sec:discussion}, we discuss the experimental results and some limitations of this study, and conclude in \secref{sec:conclusion}.
}%
{%
  これまで, 多様なロボットのナビゲーション方法が開発されてきた.
  それらは主に, Simultaneous Localization and Mapping (SLAM)や強化学習等, 事前の学習やデータ収集を必要とするものが多い.
  一方で, その適応性の高さや即応性から, Behavior-basedなロボットナビゲーションが古くから開発されてきた.
  本研究では, これらBehavior-basedなナビゲーションと現在の事前学習済み視覚-言語モデルの発展を結合することで, Open Vocabularyな反射型のナビゲーションを行うことを考える.
  一切の学習やSLAM等は行わず, 事前知識無しに, 最もシンプルな形でナビゲーションを実現する(\figref{figure:concept}).
  そこで, 全天球画像を利用する.
  全天球画像をを分割して視覚-言語モデルを適用することで, 最も指示に適切な方向に動作していくという反射型のシンプルな行動制御が可能となり, 軌道生成を含めた煩雑な探索行動の必要がなくなる.
  また, 複数の事前学習済み視覚-言語モデルを用いることで, より適切なナビゲーションが可能となることを示す.
  これまでの研究では, 事前に地図や方策を生成していたが, 本研究ではそれらは敢えて行わなず, 如何にシンプルにOpen-VocaburalyなNavigationができるかを議論する.
  各プロセス自体は目新しいものではありませんが, それを組み合わせることで, これまでにないopen-vocabularyな反射型navigationを実現しています.

  本研究の構成は以下である.
  \secref{sec:related}では, マップ生成や強化学習に基づくナビゲーション, 全天球カメラを用いたナビゲーションについて関連研究を述べる.
  \secref{sec:proposed}では, 全天球画像の展開, 視覚-言語モデルの応用方法, 言語指示に基づく反射型行動制御について述べる.
  \secref{sec:experiment}では, シンプルな定量評価実験から, より実用的な応用実験を述べる.
  \secref{sec:discussion}では, 本研究における実験結果といくつかの限界について考察し, \secref{sec:conclusion}で結論を述べる.
}%

\section{Related Works} \label{sec:related}

\subsection{Navigation with Map Construction and Reinforcement Learning}
\switchlanguage%
{%
  Most of the robot navigation methods are based on Simultaneous Localization and Mapping (SLAM) \cite{saputra2018slam, grisetti2010slam} or reinforcement learning \cite{zhu2021navigation}.
  In addition, the number of studies of open-vocabulary navigation has been increasing in recent years \cite{das2018eqa, shah2022lmnav, shafiullah2022clipfields, huang2022vlmaps, gadre2023cows, jatavallabhula2023conceptfusion}.
  Embodied Question Answering (EmbodiedQA) \cite{das2018eqa} performs reinforcement learning-based path planning in which the robot searches for answers to questions in the simulation space.
  LM-NAV \cite{shah2022lmnav} combines Large Language Model (LLM), Vision-Language Model (VLM), and Visual Navigation Model (VNM) for path planning.
  However, EmbodiedQA \cite{das2018eqa} requires reinforcement learning in simulations, and LM-NAV \cite{shah2022lmnav} requires a large number of observations of the current environment for the construction of VNM.
  The same is true for methods such as CLIP-Fields \cite{shafiullah2022clipfields}, VLMaps \cite{huang2022vlmaps}, CLIP on Wheels \cite{gadre2023cows}, and ConceptFusion \cite{jatavallabhula2023conceptfusion}, and prior learning or data collection is indispensable.
  Therefore, there are few methods that allow real-world robots to start operating immediately in an environment with no prior knowledge.
}%
{%
  Navigationの多くはSimultaneous Localization and Mapping (SLAM)を使った方法\cite{saputra2018slam, grisetti2010slam}や, 強化学習を使った方法\cite{zhu2021navigation}である.
  また, Open VocabularyによるNavigationを行った例も近年増えつつある.
  Embodied Question Answering \cite{das2018eqa}はシミュレーション空間内でロボットが質問の答えを探索する経路計画を強化学習に基づき行う.
  LM-NAV \cite{shah2022lmnav}はLarge Language Model (LLM), Vision-Language Model (VLM), Visual Navigation Model (VNM)を組み合わせ, 経路計画を行う手法を提案している.
  一方で, \cite{das2018eqa}はシミュレーション上における強化学習を行っており, \cite{shah2022lmnav}はVNMの構成のために現在の環境について大量の観測を得る必要がある.
  これらは, CLIP-Fields \cite{shafiullah2022clipfields}やVLMaps \cite{huang2022vlmaps}等の手法でも同様であり, 事前の学習やデータ収集は必須である.
  そのため, 実機ロボットが全く事前知識のない環境でも即座に動作を開始できる手法は少ない.
}%

\subsection{Navigation with Omnidirectional Camera}
\switchlanguage%
{%
  The omnidirectional camera provides a uniform view of the surroundings and can be used for various tasks.
  Kobilarov et al. \cite{kobilarov2006omnitracking} has achieved human tracking using an omnidirectional camera and Laser Range Finder (LRF), and Markovi{\'c} et al. \cite{ivan2014omnitracking} has achieved dynamic object tracking.
  Rituerto et al. \cite{rituerto2010omnislam} has developed SLAM using an omnidirectional camera, Winters et al. \cite{winters2000omnidirectional} has developed navigation using a topological map with an omnidirectional camera, and Caron et al. \cite{caron2013omniservoing} has realized visual servoing using an omnidirectional camera.
  On the other hand, these are not yet strongly linked to pretrained vision-language models.
  In this study, we apply the characteristic that the omnidirectional camera allows the robot to understand the situation all around it, eliminating the need for complicated exploratory behaviors including trajectory generation, and enabling reflex-based navigation.
  In addition, we utilize multiple pre-trained vision-language models \cite{radford2021clip, zhou2022detic} for open-vocabulary navigation.
}%
{%
  全天球画像は周りを一様に見ることができ, 多様なタスクに用いられる.
  \cite{kobilarov2006omnitracking}は全天球カメラとLaserを用いた人のトラッキング, \cite{ivan2014omnitracking}は動的物体のトラッキングを実現している.
  \cite{rituerto2010omnislam}は全天球カメラを用いたSLAMを, \cite{winters2000omnidirectional}は全天球カメラによるtopological mapを利用したNavigationを, \cite{caron2013omniservoing}はVisual Servoingを実現している.
  一方で, 未だにこれらは事前学習済み視覚-言語モデルとは大きくは結びついていない.
  本研究では, この全天球カメラにより全周の状況を把握することで軌道生成含めた煩雑な探索行動が必要なくなり, 反射型の行動が可能であるという特性を応用する.
  また, 複数の事前学習済み視覚-言語モデル\cite{radford2021clip, zhou2022detic}を用いることでOpen Vocabularyによるナビゲーションを行う.
}%

\subsection{Reflex-Based Navigation}
\switchlanguage%
{%
  In this study, ``reflex'' refers to the execution of low-level control such as direct joint angle and body velocity adjustments at a relatively fast frequency based on obtained sensory information.
  It also denotes a mechanism that directly associates sensory input with control command without requiring prior knowledge.
  Reflex-based robot navigation has been developed for a long time \cite{brooks1986subsumption, rosenblatt1989subsumption, duchon1994behaviorbased, dubrawski1994locomotion, chatterjee2001fuzzy, boyer2013underwater}.
  They have been mainly studied from the perspectives of fuzzy logic \cite{dubrawski1994locomotion, chatterjee2001fuzzy}, subsumption architecture \cite{brooks1986subsumption, rosenblatt1989subsumption}, morphological computation \cite{boyer2013underwater}, and reinforcement learning \cite{dubrawski1994locomotion}.
  On the other hand, their primary purposes are the movement to target positions represented as coordinates and collision avoidance, and open-vocabulary navigation is far from their scope.
}%
{%
  本研究で反射とは, 得られた感覚に基づき直接関節角度や速度等の低レイヤ制御が比較的速い周期で実行されることとしています.
  また, 事前知識を必要とせず, 感覚と制御入力を直接結びつけた行動を指しています.
  反射型のロボットナビゲーションは古くから開発されてきた\cite{brooks1986subsumption, rosenblatt1989subsumption, duchon1994behaviorbased, dubrawski1994locomotion, chatterjee2001fuzzy, boyer2013underwater}.
  それらは主に, fuzzy論理\cite{dubrawski1994locomotion, chatterjee2001fuzzy}, subsumption architecture \cite{brooks1986subsumption, rosenblatt1989subsumption}, mophological computation \cite{boyer2013underwater}, 強化学習\cite{dubrawski1994locomotion}の観点から研究されている.
  一方で, その目的は座標として表現された指令位置への移動と衝突回避が主である.
  そのため, 本研究で扱うOpen-VocabularyによるNavigationには程遠い.
}%

\section{Reflex-Based Open-Vocabulary Navigation Using Omnidirectional Camera and Pre-Trained Vision-Language Models} \label{sec:proposed}

\begin{figure}[t]
  \centering
  \includegraphics[width=0.7\columnwidth]{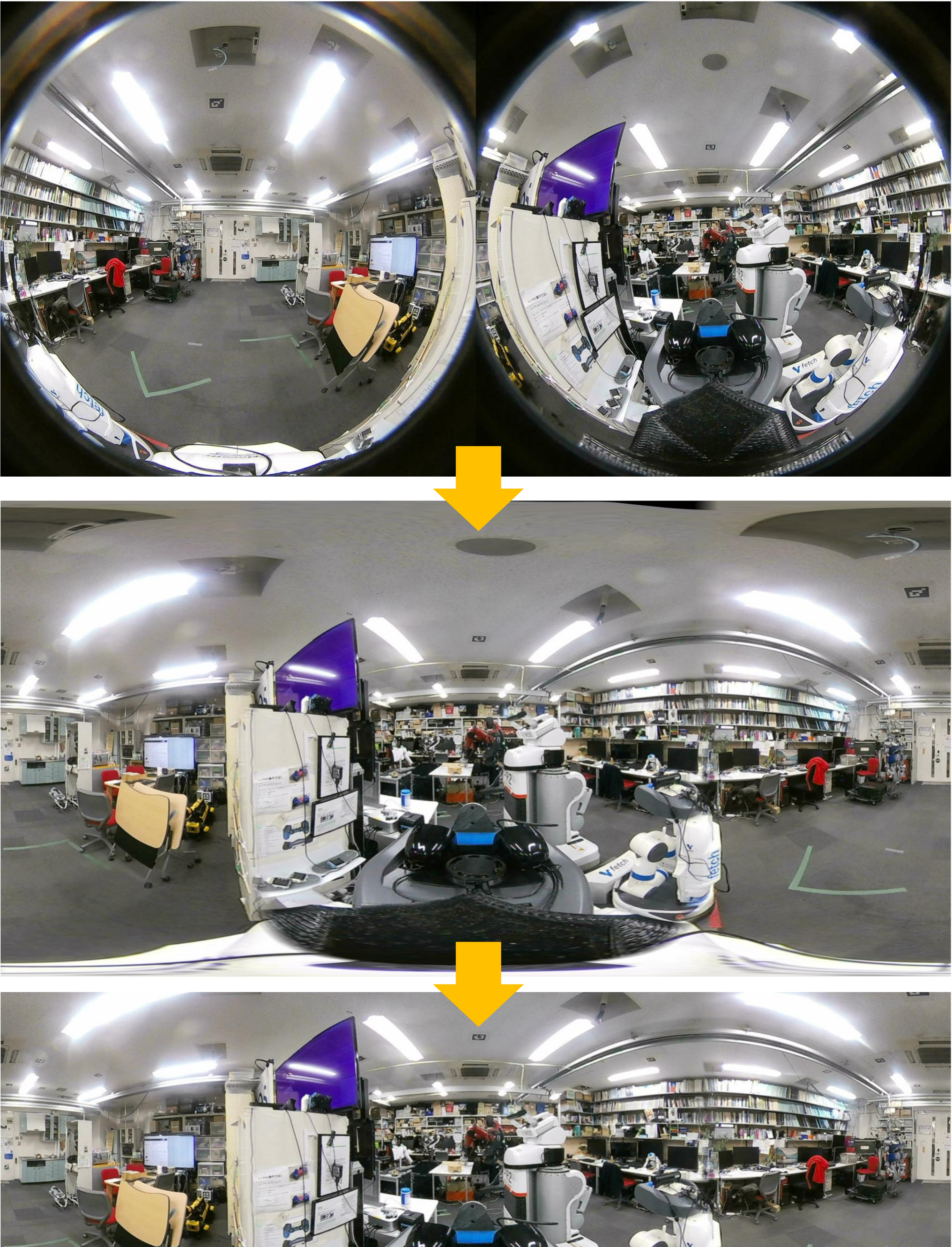}
  \caption{Dual-fisheye stitching for the expanded 360-degree omnidirectional image. The upper figure shows the fisheye images before processing, the middle figure shows the expanded image, and the lower figure shows the input image to vision-language models with unnecessary parts removed from the expanded image.}
  \label{figure:omnidirectional}
\end{figure}

\subsection{Omnidirectional Camera} \label{subsec:omnicamera}
\switchlanguage%
{%
  In this study, we use Insta360 Air Camera (Arashi Vision Inc.), which is equipped with two fisheye lenses (front and rear).
  This camera has a combination of two fisheye lenses that can provide a 360-degree field of view.
  Here, the two fisheye images are expanded and combined together \cite{ho2017fisheye}.
  This process consists of four steps.
  First, fisheye lens intensity compensation is performed.
  In order to compensate for the optical phenomenon of vignetting, the pixel intensity is modified according to the distance from the center of the lens.
  Next, fisheye unwarping is performed.
  In order to produce a natural photographic image, each pixel is expanded by a geometric transformation in the order of unit sphere and square image.
  Then, the geometric misalignment between the two images is minimized from manual annotation in the overlapping regions.
  Finally, the two images are blended.
  The fisheye images before and after these steps are shown in \figref{figure:omnidirectional}.
  There are many unnecessary areas such as the ceiling and the robot's head, at the top and bottom of the 2000$\times$1000 expanded image.
  If this image is input to the vision-language models as it is, the information will become biased.
  Therefore, in practical use, we remove the lower and upper areas of the image, making it a 2000$\times$500 image $V$ as shown in the lower figure of \figref{figure:omnidirectional}.
}%
{%
  本研究では全天球カメラとして, 魚眼レンズを前後に2つ備えたInsta360 Air Camera (Arashi Vision Inc.)を用いる.
  単体の魚眼レンズではなく, それを複数合わせることで360度の視野に対応することができる.
  本研究では得られた2つの魚眼画像を展開, 合成して用いる\cite{ho2017fisheye}.
  これは主に4つのプロセスからなる.
  まず, fisheye lens intensity compensationを行う.
  光学現象であるvignettingを補償するため, レンズの中心からの距離に応じてpixel intensityを修正する.
  次に, fisheye unwarpingを行う.
  写真のように自然な画像を生成するために, それぞれの画素を単位球, 正方画像の順に幾何的な変換を行い展開する.
  次に, overlapping regionsにおける手動のannotationから2つの画像のgeometric misalignmentを最小化する.
  最後に, これら2つの画像のblendingを行う.
  これらの手順を行う前の魚眼画像と, 行った後の展開画像を\figref{figure:omnidirectional}に示す.
  2000$\times$1000の展開画像の上下にはロボットの頭や天井等の無駄な部分が多く, これらをそのまま視覚-言語モデルに入力すると情報が偏る.
  そのため, 実際に使用する際は下と上の部分を削除し, \figref{figure:omnidirectional}の下図に示すような2000$\times$500の画像$V$に成形して用いている.
}%

\begin{figure}[htb]
  \centering
  \includegraphics[width=1.0\columnwidth]{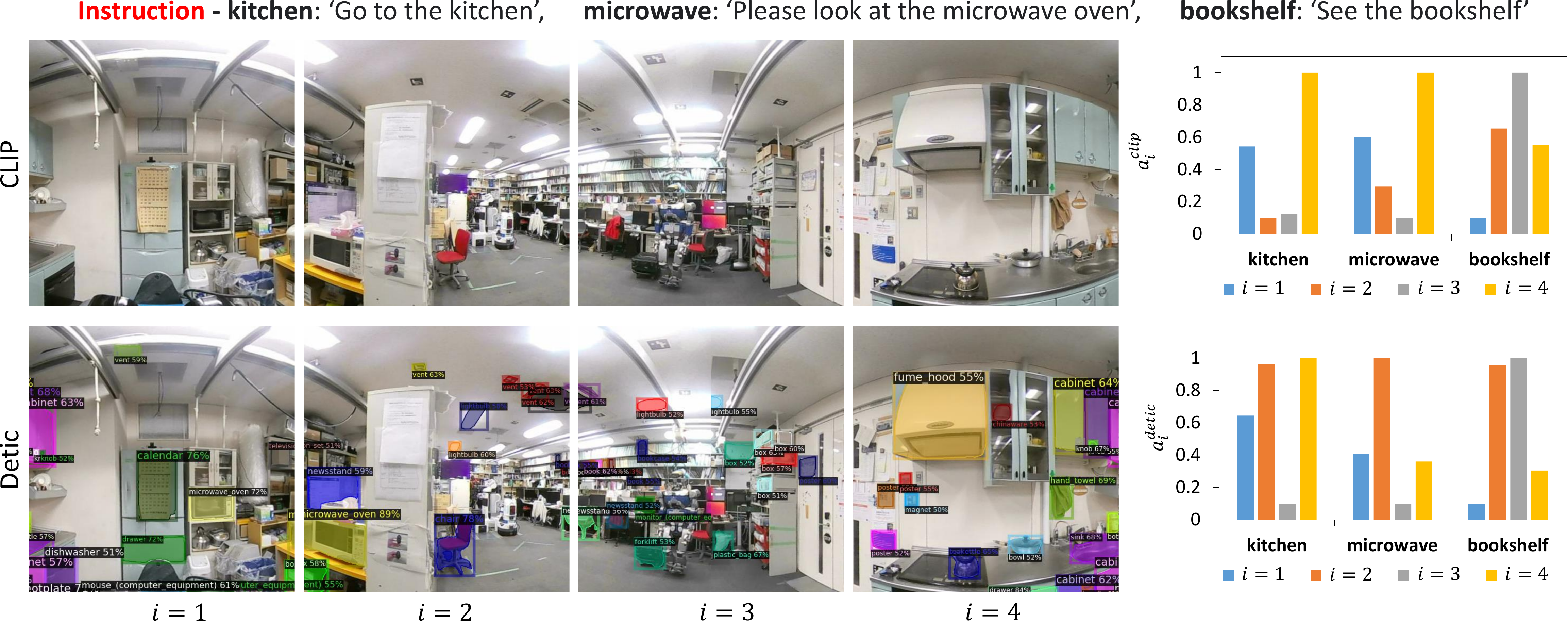}
  \caption{The preliminary experiments of using large-scale vision-language models for open-vocabulary navigation. The left figure shows the split image and the recognition result, and the right graph shows the average of the transformed similarity $a$ for 10 repetitions of each instruction for CLIP and Detic: \textbf{kitchen} - ``Go to the kitchen'', \textbf{microwave} - ``Please look at the microwave oven'', \textbf{bookshelf} - ``See the bookshelf''.}
  \label{figure:vlm}
\end{figure}

\subsection{Application of Pre-Trained Vision-Language Models} \label{subsec:vlm}
\switchlanguage%
{%
  In this study, we use CLIP \cite{radford2021clip} and Detic \cite{zhou2022detic} as pre-trained vision-language models.
  CLIP is a method to transform images and languages into vectors in the same embedding space, and Detic is a recognizer capable of detecting various classes of objects from images.
  Current VLMs are capable of four tasks: Generation Task, Understanding Task, Retrieval Task, and Grounding Task \cite{li2022largemodels}.
  Among these, models like OFA \cite{wang2022ofa} and BLIP2 \cite{li2023blip2}, which are capable of Generation Task and Understanding Task, are relatively heavy in processing and not suitable for reflex-based control.
  On the other hand, Retrieval Task and Grounding Task have low computational cost and are suitable for this study.
  Models capable of Retrieval Task include CLIP \cite{radford2021clip} and ImageBind \cite{girdhar2023imagebind}, while models capable of Grounding Task include Detic \cite{zhou2022detic} and ViLD \cite{gu2022vild}.
  Amog these models, we selected the representative models CLIP and Detic.
  Note that, for Detic, we use the object classes in LVIS \cite{gupta2019lvis} as objects to be detected.

  First, the image obtained by \secref{subsec:omnicamera} is split into $N_{split}$ pieces.
  Here, there can be overlaps between the pieces.
  Next, CLIP and Detic are applied to the obtained split images $V_i$ ($1 \leq i \leq N_{split}$).
  Let $\bm{v}^{clip}_{i}$ be a vector obtained by applying CLIP to $V_i$.
  From Detic, labels and bounding boxes of various objects are obtained.
  The detected labels are sorted in descending order by the size of the bounding box, and are separated by commas to form a sentence (e.g. ``table, monitor, monitor, monitor, apple, knife'').
  This sentence is transformed into a vector $\bm{v}^{detic}_{i}$ by Sentence-BERT \cite{reimers2019sentencebert}.
  In practice, Detic is applied not to the split image $V_i$ but to the expanded image $V$ from the viewpoint of computation time, and the obtained result is split.

  When the robot receives a linguistic instruction $Q$, it is transformed into a vector $\bm{q}^{clip}$ by CLIP and into a vector $\bm{q}^{detic}$ by Sentence-BERT.
  Assuming that the obtained vectors are all normalized, we obtain the following cosine similarity $s$ and the transformed similarity $a$,
  \begin{align}
    s^{clip}_{i} &= \bm{v}^{clip}_{i}\cdot\bm{q}^{clip}\\
    a^{clip}_{i} &= 0.1 + 0.9\frac{s^{clip}_{i}-s^{clip}_{min}}{s^{clip}_{max}-s^{clip}_{min}}\\
    s^{detic}_{i} &= \bm{v}^{detic}_{i}\cdot\bm{q}^{detic}\\
    a^{detic}_{i} &= 0.1 + 0.9\frac{s^{detic}_{i}-s^{detic}_{min}}{s^{detic}_{max}-s^{detic}_{min}}
  \end{align}
  where $s^{\{clip, detic\}}_{\{min, max\}}$ denotes the minimum and maximum values of $s^{\{clip, detic\}}_{i}$ ($1 \leq i \leq N_{split}$).
  The minimum value of $a$ is converted to 0.1 and the maximum value of $a$ is converted to 1.0 in order to measure $s^{clip}_{i}$ and $s^{detic}_{i}$ on the same scale.
  Note that the minimum value is set to 0.1 instead of 0 because of the way it is used in \secref{subsec:mapping}.

  As an example, the result for $N_{split}=4$ is shown in \figref{figure:vlm}.
  The following three instructions are given: \textbf{kitchen} - ``Go to the kitchen'', \textbf{microwave} - ``Please look at the microwave oven'', \textbf{bookshelf} - ``See the bookshelf''.
  The right figure of \figref{figure:vlm} shows the average of $a_i$ for 10 repetitions of each instruction when $i=\{1, 2, 3, 4\}$.
  First, in the case of \textbf{kitchen}, the images of $i=\{1, 4\}$ match the instruction.
  $a_{\{1, 4\}}$ are high for CLIP as intended because it judges the similarity between the instruction and the image by looking at the image as a whole.
  On the other hand, Detic recognizes each object in the image separately.
  Therefore, Detic recognized the microwave oven, an object that is usually located in the kitchen, making $a_2$ as high as $a_4$.
  Next, in the case of \textbf{microwave}, the microwave oven is actually in the image when $i=\{1, 2\}$.
  $a_2$ is high in Detic as intended because it recognizes individual objects.
  Note that $a_1$ is not so high because the bounding box of the microwave oven is small.
  On the other hand, CLIP sees the image as a whole and it misses individual objects, making $a_2$ low.
  Since the microwave oven is usually located in the kitchen, we can see that $a_i$ in CLIP shows similar distributions in the cases of \textbf{microwave} and \textbf{kitchen}.
  Finally, in the case of \textbf{bookshelf}, the bookshelf is actually in the image when $i=\{2, 3\}$, and we can see that $a_{\{2, 3\}}$ are high as intended for both models.
  Thus, depending on the vision-language model used, there are instructions that are good or bad at being recognized.
}%
{%
  本研究では事前学習済み視覚-言語モデルとして, CLIP \cite{radford2021clip}とDetic \cite{zhou2022detic}を用いる.
  CLIPは画像や言語を同じ埋め込み空間のベクトルに変換する手法であり, Deticは画像から多様なクラスの物体検出が可能な認識器である.
  現在のVLMが可能なタスクには, Generation Task, Understanding Task, Retrieval Task, Grounding Taskの4つが存在している\cite{li2022largemodels}.
  その中でも, Generation TaskやUnderstanding Taskが可能なOFA \cite{wang2022ofa}やBLIP2 \cite{li2023blip2}は比較的処理が重く, 反射型の制御には向かない.
  Retrieval Taskが可能なモデルはCLIP \cite{radford2021clip}やImageBind \cite{girdhar2023imagebind}, Grounding Taskが可能なモデルはDetic \cite{zhou2022detic}やViLD \cite{gu2022vild}が存在している.
  本研究ではその中でも代表的なCLIPとDeticを選択した.
  なお, Deticについては検出する物体としてLVIS \cite{gupta2019lvis}における物体クラスを用いている.

  まず, \secref{subsec:omnicamera}で得られた画像を$N_{split}$個に分割する.
  この際, それぞれの画像にオーバーラップがあっても良い.
  次に, 得られた分割画像$V_i$ ($1 \leq i \leq N_{split}$)に対してそれぞれCLIPとDeticを適用する.
  ここで, $V_i$に対してCLIPを適用することで得られたベクトルを$\bm{v}^{clip}_{i}$とする.
  また, Deticからは多様な物体のラベルとバウンディングボックスが得られる.
  この検出されたラベルをバウンディングボックスの大きさで降順にソートし, カンマで区切って繋げ一文とする(e.g. ``table, monitor, monitor, monitor, apple, knife'').
  この一文をSentence-BERT \cite{reimers2019sentencebert}によってベクトル$\bm{v}^{detic}_{i}$に変換する.
  なお, 実際には計算時間の観点から, Deticは分割画像$V_i$ではなく展開画像$V$に適用した結果を, 分割して用いている.

  言語による指示$Q$をロボットが受け取った際, これをCLIPによりベクトル$\bm{q}^{clip}$に, Sentence-BERTによりベクトル$\bm{q}^{detic}$に変換する.
  得られたベクトルが全て正規化されているとして, 以下のコサイン類似度$s$, これを変換した類似度$a$を得る.
  \begin{align}
    s^{clip}_{i} &= \bm{v}^{clip}_{i}\cdot\bm{q}^{clip}\\
    a^{clip}_{i} &= 0.1 + 0.9\frac{s^{clip}_{i}-s^{clip}_{min}}{s^{clip}_{max}-s^{clip}_{min}}\\
    s^{detic}_{i} &= \bm{v}^{detic}_{i}\cdot\bm{q}^{detic}\\
    a^{detic}_{i} &= 0.1 + 0.9\frac{s^{detic}_{i}-s^{detic}_{min}}{s^{detic}_{max}-s^{detic}_{min}}
  \end{align}
  ここで, $s^{\{clip, detic\}}_{\{min, max\}}$は, $s^{\{clip, detic\}}_{i}$ ($1 \leq i \leq N_{split}$)の最小値と最大値を表す.
  $a$は$s^{clip}_{i}$と$s^{detic}_{i}$を同じ尺度で測るため, 最小値を0.1, 最大値を1として変換したものである.
  なお, \secref{subsec:mapping}での利用方法の関係から, 最小値を0ではなく0.1としている.

  例として, $N_{split}=4$のときの結果を\figref{figure:vlm}に示す.
  指示として, \textbf{kitchen}: ``Go to the kitchen'', \textbf{microwave}: ``Please look at the microwave oven'', \textbf{bookshelf}: ``See the bookshelf''の3つを与えている.
  \figref{figure:vlm}の右図に, $i=\{1, 2, 3, 4\}$のときの, それぞれの指示における10ステップ分の$a_i$の平均が示されている.
  まず\textbf{kitchen}のケースでは, 主に$i=\{1, 4\}$が正しいと推測されるべきである.
  CLIPは画像全体を見て近さを判断するため正しく$a_{\{1, 4\}}$が高いが, Deticでは$a_2$も$a_4$と同様に高い.
  これは, Deticが物体を個別に見るため, $i=2$についてはmicrowave ovenという大抵はkitchenにある物体を認識してしまったからであると考えられる.
  次に\textbf{microwave}のケースでは, $i=\{1, 2\}$において実際にmicrowave ovenが画像に映っている.
  CLIPは画像全体を見てしまうため, \textbf{kitchen}のケースとは逆に個別の物体を見落としてしまい$a_2$が低い.
  microwave ovenは大抵kitchenにあるため, CLIPにおける$a_i$は\textbf{microwave}と\textbf{kitchen}のケースで似た分布を示していることが分かる.
  一方Deticの場合は個別の物体を認識するため, 正しく$a_2$が高くなっている.
  なお, $i=1$のmicrowaveはバウンディングボックスが小さいため, $a_1$はあまり高くない.
  最後に\textbf{bookshelf}のケースについては, 本棚の面積が$i=3$, $i=2$の順で大きいが, 両者とも正しく$a_3$, $a_2$が高くなっていることがわかる.
  このように, 視覚-言語モデルはその利用方法によって認識が得意・不得意な指示が存在している.
}%

\begin{figure}[t]
  \centering
  \includegraphics[width=0.4\columnwidth]{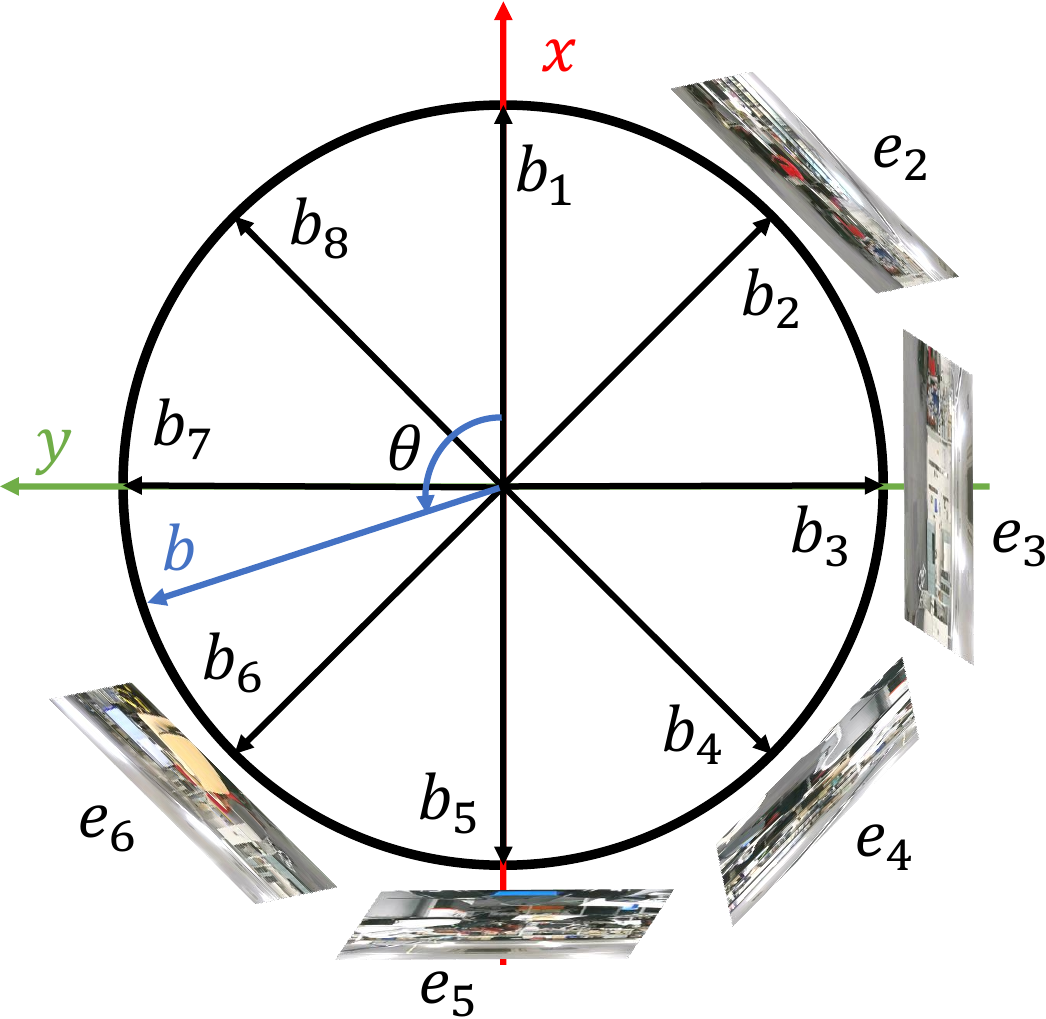}
  \caption{The configuration when mapping linguistic instructions to robot wheeled-base motion.}
  \label{figure:mapping}
\end{figure}

\subsection{Mapping from Linguistic Instruction to Motion} \label{subsec:mapping}
\switchlanguage%
{%
  We map $a^{\{clip, detic\}}_{i}$ obtained in \secref{subsec:vlm} to the actual robot motion (\figref{figure:mapping}).
  First, we calculate the following evaluation value $e_{i}$ in order to resolve the problem that $a^{clip}_{i}$ and $a^{detic}_{i}$ each have its own strengths and weaknesses as described in \secref{subsec:vlm}.
  \begin{align}
    e_{i} = a^{clip}_{i}a^{detic}_{i}
  \end{align}
  The minimum value of $a_i$ is set to 0.1 as described in \secref{subsec:vlm} to ensure that $e_i$ does not become 0 even when either $s^{clip}_{i}$ or $s^{detic}_{i}$ is very small.
  By using such metrics, we accommodate scenarios where one of $s^{\{clip, detic\}}_{i}$ is small while the other is large.
  Next, we retrieve $N_{extract}$ number of $e_{i}$ in descending order.
  Here, let $\bm{b}_{i}$ be the unit direction vector toward which the center point of the image $V_{i}$ faces, and let $i_{j}$ be $i$ of the $j$-th largest $e_{i}$ ($1 \leq j \leq N_{extract}$).
  The direction vector $\bm{b}$ that the robot should move toward is calculated as follows.
  \begin{align}
    \bm{b} = \Sigma_{j} (N_{extract}+1-j)\bm{b}_{i_{j}} / \Sigma_{j} (N_{extract}+1-j)
  \end{align}
  This calculation is adopted to make a proper judgement based on the totality of the surrounding circumstances and to avoid the problem of falling into a local minima by making the robot move in the direction with the largest $e_{i}$.
  We map this $\bm{b}$ to the robot motion.
  For an omnidirectional wheeled robot with omni wheels or mecanum wheels, it is sufficient to simply generate a translation speed in the direction of $\bm{b}$.
  On the other hand, for a two-wheeled robot, the mapping is as follows,
  \begin{align}
    a_{linear} &= 1.0 \;\;\;\text{if $|\theta| < C_{thre}$}\\
    a_{rotate} &= k\theta
  \end{align}
  where $a_{\{linear, rotate\}}$ is the velocity of the robot in the translational and rotational direction (\{m/s, rad/s\}), $\theta$ is the angle between $\begin{pmatrix}1&0\end{pmatrix}^{T}$ and $\bm{b}$ and is expressed from $-\pi$ to $\pi$, $k$ is a proportionality constant, and $C_{thre}$ is a threshold value.
  Note that the rotational directions of $i$ and $\theta$ are different because the coordinate systems of the robot and the image are different.
}%
{%
  \secref{subsec:vlm}で得られた$a^{\{clip, detic\}}_{i}$を実際のロボットの動きにマッピングする(\figref{figure:mapping}).
  まず, \secref{subsec:vlm}で述べたような, $a^{clip}_{i}$と$a^{detic}_{i}$それぞれの得意と不得意の問題を解消するため, 以下の指標$e_{i}$を計算する.
  \begin{align}
    e_{i} = a^{clip}_{i}a^{detic}_{i}
  \end{align}
  $a_i$の最小値を0.1としたのは, $s^{clip}_{i}$または$s^{detic}_{i}$が0のときでも, $e_i$が0にならないようにするためである.
  $s^{\{clip, detic\}}$の一方が小さく, もう一方が大きい場合も考慮するため, このような指標を用いている.
  次に, 画像$V_{i}$の中心点が向く単位方向ベクトルを$\bm{b}_{i}$とする.
  ここで, $e_{i}$を大きいものから順に$N_{extract}$個取り出す.
  $j$番目に$e_{i}$の大きな$i$を$i_{j}$としたとき($1 \leq j \leq N_{extract}$), ロボットが動作すべき方向ベクトル$\bm{b}$を以下のように計算する.
  \begin{align}
    \bm{b} = \Sigma_{j} (N_{extract}+1-j)\bm{b}_{i_{j}} / \Sigma_{j} (N_{extract}+1-j)
  \end{align}
  一番$e_{i}$の大きな方向に動作すると, 適切に周囲の状況を総合して判断できなくなり, 局所解に陥ることがあったため, このような方式としている.
  この$\bm{b}$をロボットの動きにマッピングする.
  オムニホイールやメカナムホイールを用いた全方位台車型ロボットであれば, その方向に単順に並進速度を出せば良い.
  一方で二輪台車の場合は以下のようにマッピングする.
  \begin{align}
    a_{linear} &= 1.0 \;\;\;\text{if $|\theta| < C_{thre}$}\\
    a_{rotate} &= k\theta
  \end{align}
  ここで, $a_{\{linear, rotate\}}$はそれぞれ台車の並進・回転方向の速度であり(\{m/s, rad/s\}), $\theta$は$\begin{pmatrix}1&0\end{pmatrix}^{T}$と$\bm{b}$の間の角度を$-\pi$から$\pi$の間で表現した値, $k$は比例係数, $C_{thre}$は定数を表す.
  なお, ロボットの座標系と画像の座標系が異なるため, $i$と$\theta$の回転方向は異なっている.
}%

\subsection{Other Settings} \label{subsec:others}
\switchlanguage%
{%
  In this study, we conduct experiments using the mobile robot Fetch.
  Since the robot is equipped with LRF, it can be constrained not to move in the direction of obstacles.
  Of course, it is possible to provide the robot with a bumper and a contact sensor, and to construct another reflective behavior for obstacle avoidance.
  By incorporating obstacle avoidance, the robot can stop appropriately without going too far in the target direction.
}%
{%
  本研究ではモバイルロボットFetchを用いて実験を行う.
  Laser Range Finderを搭載しているため, 障害物の方向には速度が出ないように制約をかけている.
  もちろんロボットにバンパーと接触センサを用意して, 障害物回避のための別の反射回路を組むことも可能である.
  障害物回避を組み込むことで, ロボットが進み過ぎずに適切に止まることができる.
}%

\begin{figure}[t]
  \centering
  \includegraphics[width=0.8\columnwidth]{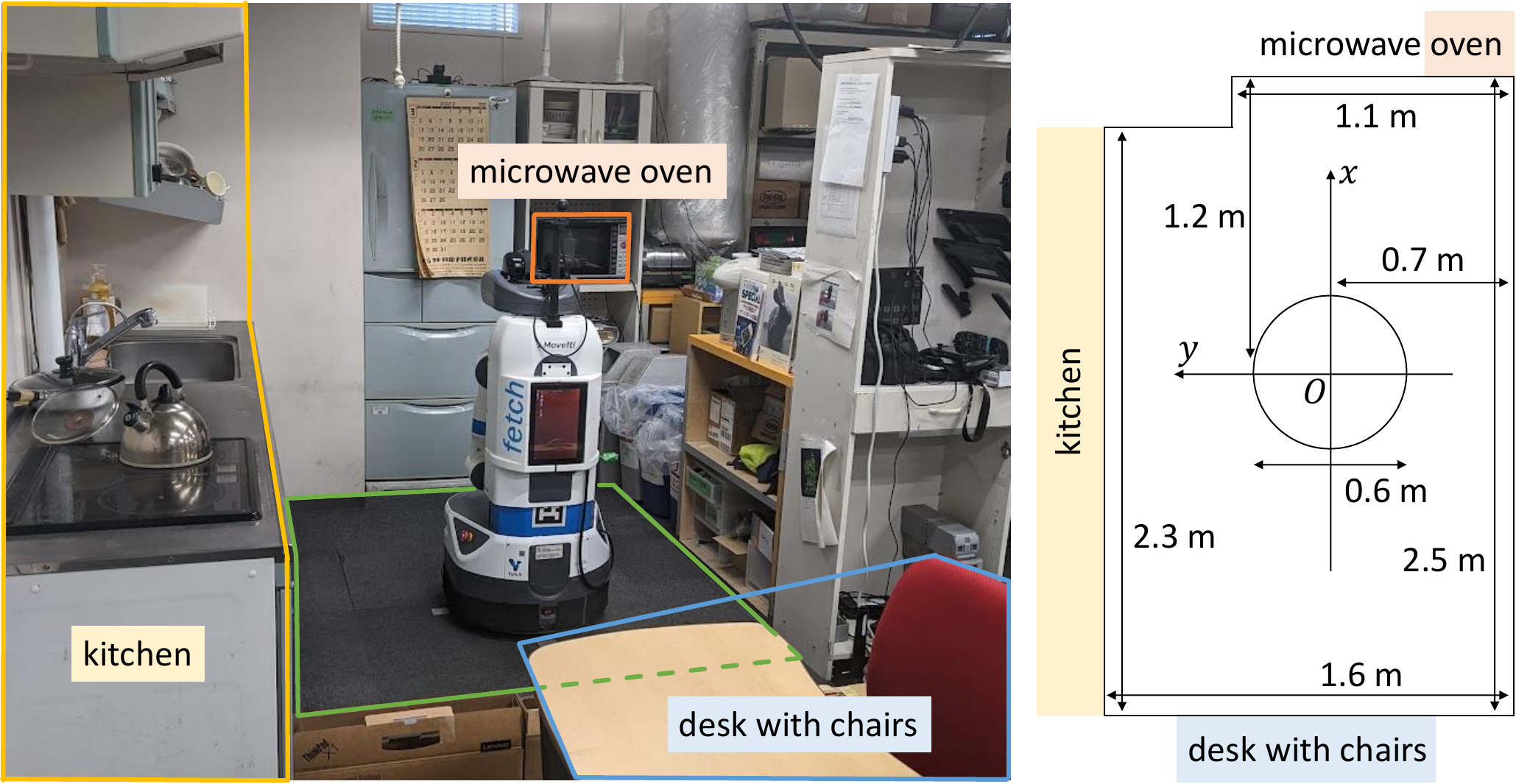}
  \caption{The environmental setup of the basic experiment. The mobile robot Fetch is placed in a small area surrounded by the kitchen, microwave oven, and desk with chairs.}
  \label{figure:basic-setup}
\end{figure}

\begin{figure}[htb]
  \centering
  \includegraphics[width=1.0\columnwidth]{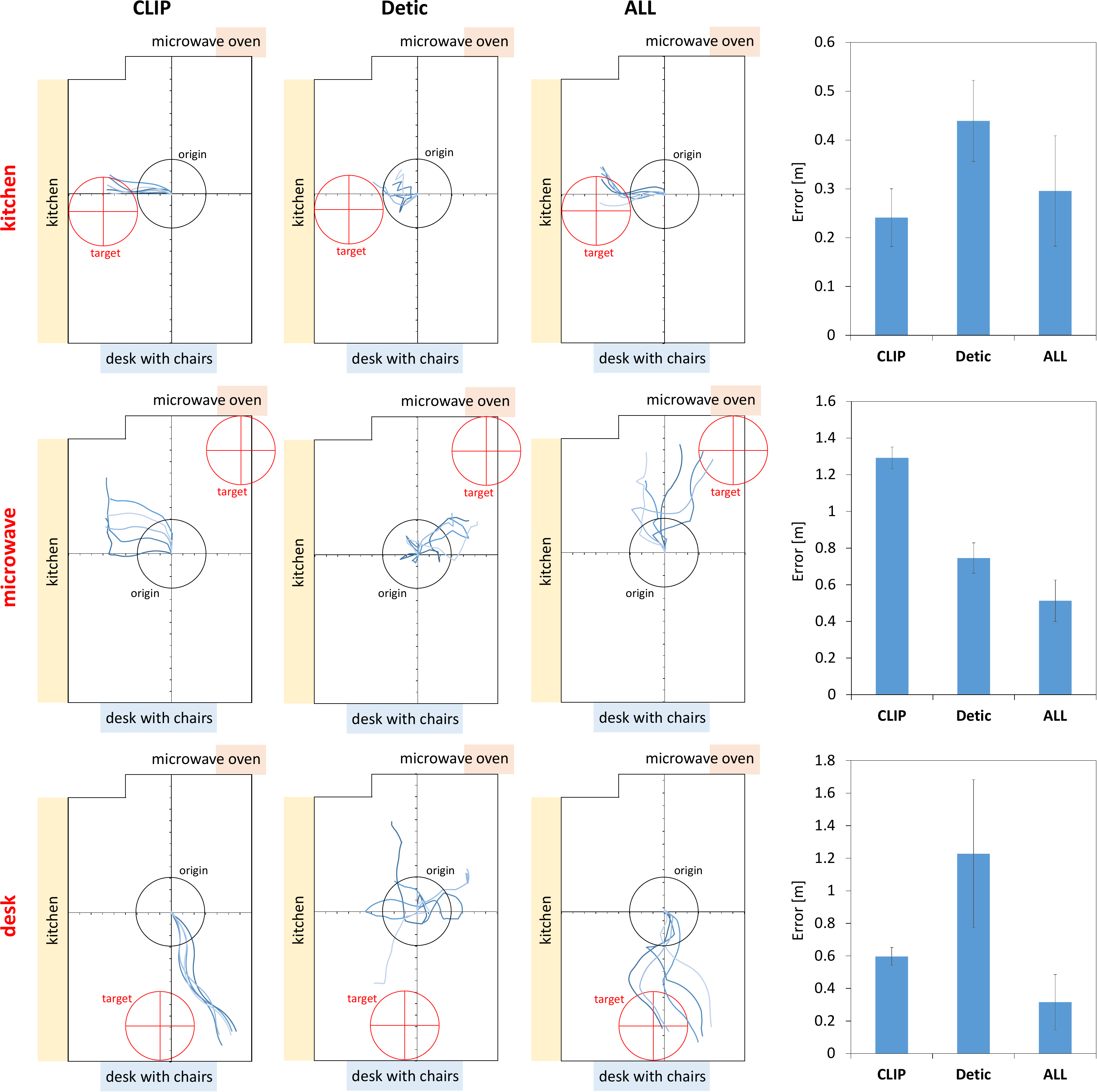}
  \caption{The trajectories of the mobile robot Fetch and the error between the robot's final and target positions in the basic experiment. We prepared three instructions: \textbf{kitchen} - ``Go to the kitchen'', \textbf{microwave} - ``Please look at the microwave oven'', and \textbf{desk} - ``See the desk with chairs'', and compared the proposed method \textbf{ALL} with \textbf{CLIP} and \textbf{Detic}.}
  \label{figure:basic-exp}
\end{figure}

\section{Experiments} \label{sec:experiment}
\switchlanguage%
{%
  In this study, we first demonstrate the performance of our method quantitatively in a small area surrounded on all sides.
  Then, we show a series of navigation experiments using our method in a larger space.
  In this study, we set $N_{split}=8$, $C_{extract}=2$, $C_{thre}=0.6$ [rad], and $k=0.5$, and the control of \secref{subsec:mapping} is operated at 10 Hz.
  With $N_{split}=8$, inference is performed at a maximum of 12 Hz for CLIP and 5 Hz for Detic on a machine equipped with an i7-6850K CPU and an Nvidia GeForce RTX 3090 GPU.
  While inference time for CLIP increases proportionally to $N_{split}$, inference time for Detic remains independent of $N_{split}$ as it is directly applied to the expanded image $V$.
}%
{%
  本研究ではまず四方を囲われた箱庭の中で本研究の性能を定量的に示す.
  その後, より広い空間における本手法を用いた一連のnavigation実験を示す.
  本研究では$N_{split}=8$, $C_{extract}=2$, $C_{thre}=0.6$ [rad], $k=0.5$とし, \secref{subsec:mapping}の制御は10 Hzで動作させる.
  また, $N_{split}=8$で, CPUがi7-6850K, GPUがNvidia GeForce RTX 3090のマシンにおいて, CLIPが最大12 Hz, Deticが最大5Hzで推論可能である.
  なお, CLIPについては分割数に比例して推論時間が増加しますが, Deticについては展開画像$V$に直接適用するため, 推論時間は分割数に依存しません.
}%

\subsection{Basic Experiment} \label{subsec:basic-exp}
\switchlanguage%
{%
  The environmental setup for the basic experiment is shown in \figref{figure:basic-setup}.
  An area is surrounded by shelves and walls on all four sides and is about 2.5$\times$1.6 m in size.
  The footprint size of Fetch is about 0.6$\times$0.6 m.
  In this experiment, we prepared three instructions: \textbf{kitchen} - ``Go to the kitchen'', \textbf{microwave} - ``Please look at the microwave oven'', and \textbf{desk} - ``See the desk with chairs''.
  The kitchen, microwave oven, and desk with chairs are arranged as in \figref{figure:basic-setup}.
  In this experiment, we compare three methods: the proposed method \textbf{ALL}, \textbf{CLIP} using only CLIP with $e_{i} = a^{clip}_{i}$, and \textbf{Detic} using only Detic with $e_{i} = a^{detic}_{i}$.

  The results of five navigation experiments for each instruction using each of the three methods are shown in \figref{figure:basic-exp}.
  The left figures show the trajectories of Fetch for each of the five navigation experiments.
  Note that Fetch is stopped when it touches the walls/shelves, or once 30 seconds have passed since the start of the experiment.
  Ideally, the robot should move toward ``target'' in \figref{figure:basic-exp}, which is the center of the position indicated by the instruction, with ``origin'' as the initial position.
  However, in some cases, the error does not need to be zero because ``kitchen'' or ``desk'' can refer to a large area, but the error should be as small as possible.
  The right figure of \figref{figure:basic-exp} shows the average and variance of the error between the robot's final position and the target position.

  For \textbf{kitchen}, we can see that the robot moves correctly toward the kitchen when using the methods of \textbf{CLIP} and \textbf{ALL}.
  On the other hand, for \textbf{Detic}, the robot meanders and does not move in the expected direction.
  The final position error of \textbf{Detic} is larger than those of \textbf{CLIP} and \textbf{ALL}.
  Although the error of \textbf{CLIP} is slightly smaller than that of \textbf{ALL}, the performance is considered to be almost the same because of the wider area of the kitchen as mentioned above.
  For \textbf{microwave}, we can see that the robot moves correctly toward the microwave oven in the case of \textbf{ALL}.
  In the case of \textbf{Detic}, the robot moves toward the microwave oven to some extent, but its performance is not as good as that of \textbf{ALL} due to meandering.
  On the other hand, in the case of \textbf{CLIP}, the robot moves toward the kitchen and its performance is significantly worse.
  For \textbf{desk}, the performance is high in the order of \textbf{ALL}, \textbf{CLIP}, and \textbf{Detic}.
  It can be seen that the performance of \textbf{ALL} is the best throughout all the experiments.
}%
{%
  基礎実験を行う環境セットアップを\figref{figure:basic-setup}に示す.
  四方が棚や壁で囲まれており, 大きさは約2.5$\times$1.6 mである.
  また, Fetchのフットプリントは約0.6$\times$0.6 mである.
  本実験では, 3つの指示\textbf{kitchen}: ``Go to the kitchen'', \textbf{microwave}: ``Please look at the microwave oven'', \textbf{desk}: ``See the desk with chairs''によりロボットを動作させる.
  そのため, \figref{figure:basic-setup}にはkitchen, microwave oven, desk with chairsが存在している(\secref{subsec:vlm}における片方のmicrowave ovenは隠した).
  また, 本実験では提案手法を\textbf{ALL}として, $e_{i} = a^{clip}_{i}$としてCLIPのみを用いた\textbf{CLIP}, $e_{i} = a^{detic}_{i}$としてDeticのみを用いた\textbf{Detic}も合わせた3種類の手法を比較する.

  それぞれの指示について, 3種類の手法を用いたnavigation実験を5回ずつ行った結果を\figref{figure:basic-exp}に示す.
  左図はそれぞれ実験5回分のFetchの軌道を示している.
  なお, 動かない, または30 sec経ったら動作を停止している.
  ``origin''を初期位置として, 指示位置の中心である``target''に向かうのが理想であるが, 例えば``kitchen''は広いため, 誤差が0になる必要はない.
  右図はロボットの最終到達位置と指令位置の間の誤差を示している.

  まず\textbf{kitchen}について, \textbf{CLIP}と\textbf{ALL}の手法では同様にkitchenの方にロボットが正しく移動していることがわかる.
  一方, \textbf{Detic}ではロボットが蛇行しており思ったとおりの方向に進めていない.
  最終的な位置誤差についても, \textbf{Detic}が\textbf{CLIP}や\textbf{ALL}に比べると大きいことがわかる.
  なお, その誤差は\textbf{CLIP}の方が\textbf{ALL}よりも多少小さいが, 前述の通りkitchenの範囲が広いため, 性能はほとんど同様だと考えられる.
  次に\textbf{microwave}について, \textbf{ALL}の手法ではmicrowave ovenの方にロボットが正しく移動していることがわかる.
  \textbf{Detic}の場合は, ある程度microwave ovenの方に移動してはいるが, 蛇行しており性能は\textbf{ALL}ほど高くない.
  一方, \textbf{CLIP}の場合はkitchenの方へ進んでしまっており, その性能は著しく悪い.
  最後に, \textbf{desk}について, その性能は\textbf{ALL}, \textbf{CLIP}, \textbf{Detic}の順に高い.
  全体を通して\textbf{ALL}の性能が最も高いことが見て取れる.
}%

\begin{figure}[t]
  \centering
  \includegraphics[width=1.0\columnwidth]{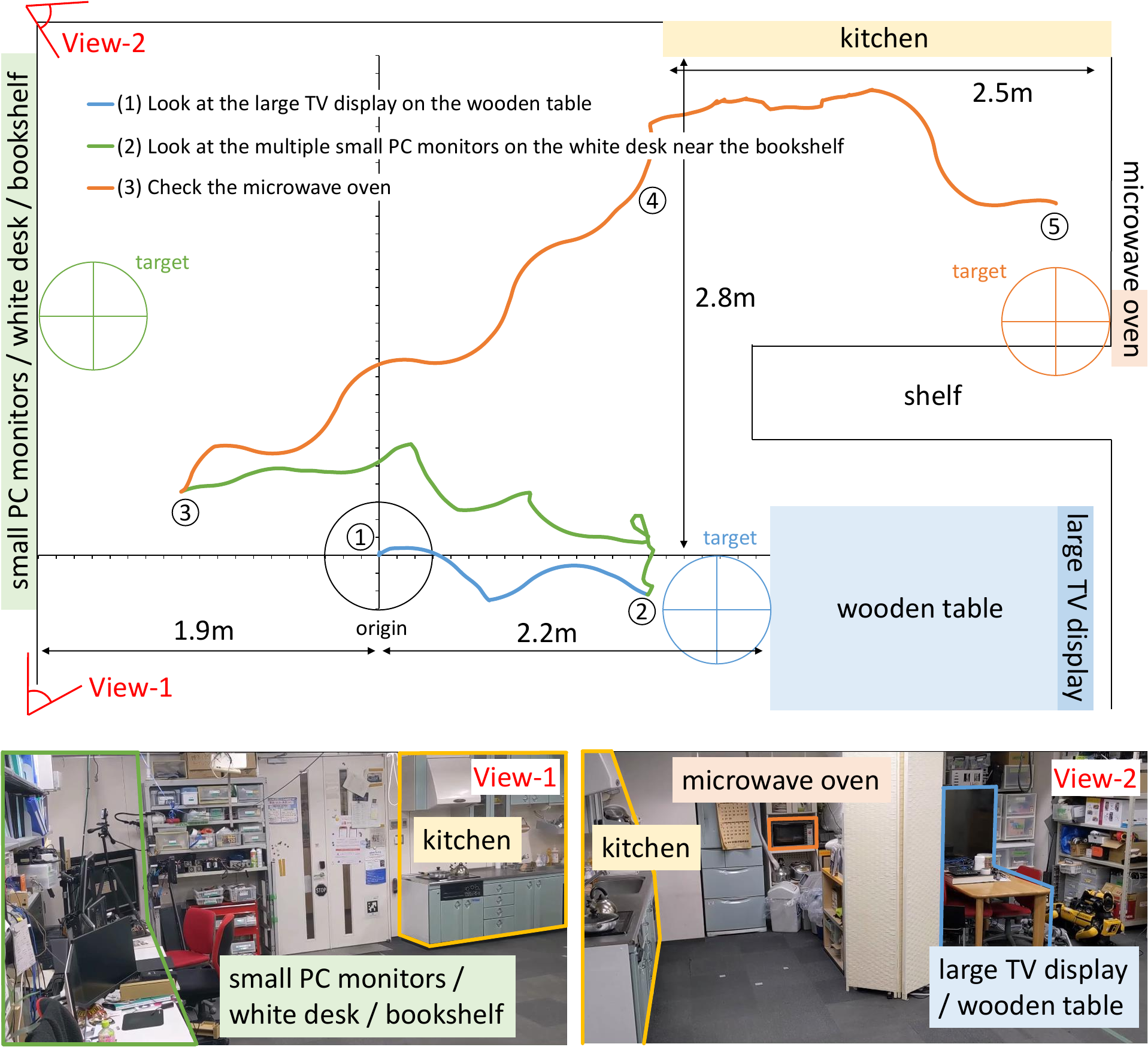}
  \caption{The trajectory of the mobile robot Fetch in the advanced navigation experiment.}
  \label{figure:advanced-traj}
\end{figure}

\begin{figure}[t]
  \centering
  \includegraphics[width=1.0\columnwidth]{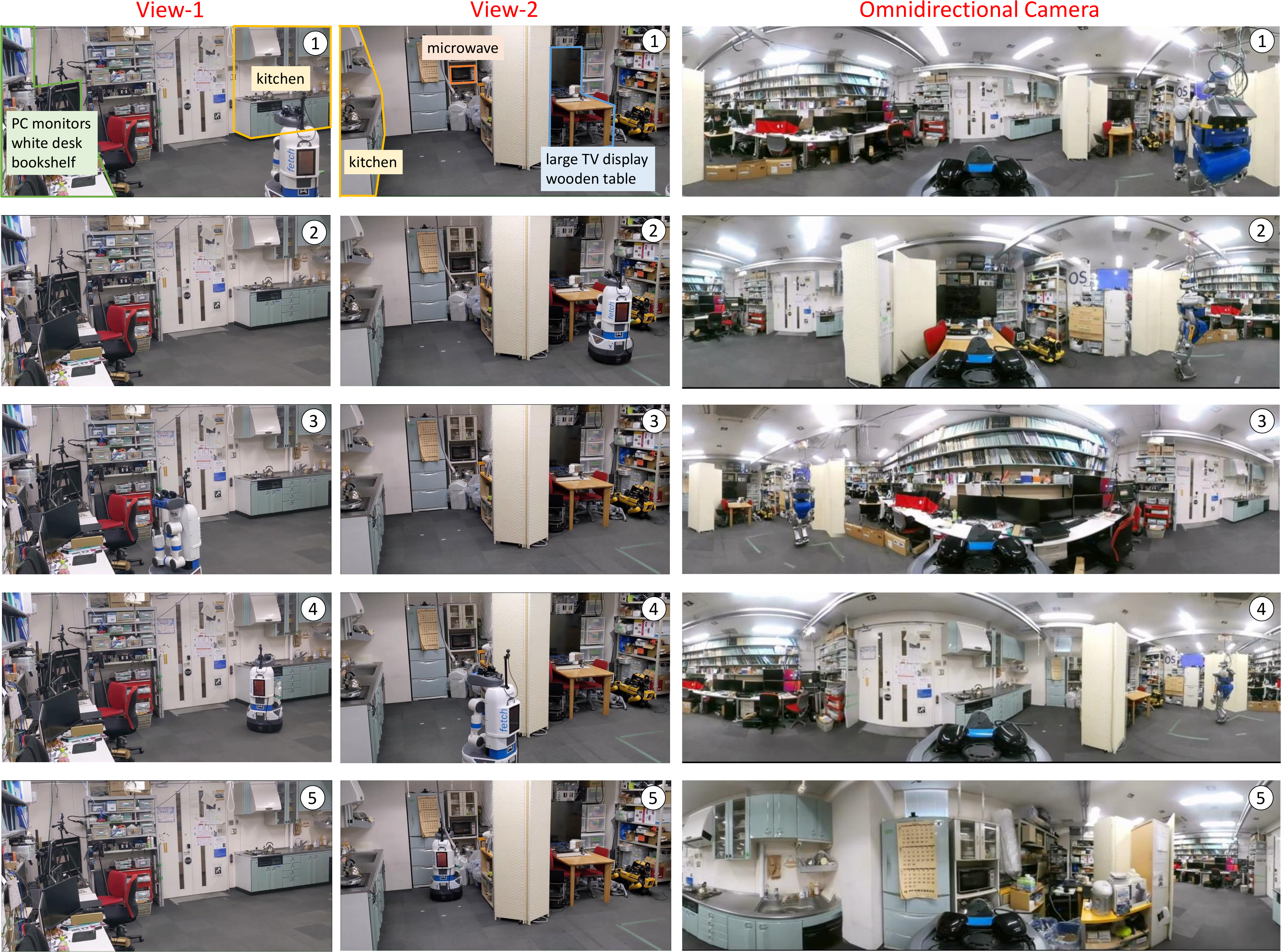}
  \caption{The advanced navigation experiment. The instruction is continuously changed in the following order: (1) ``Look at the large TV display on the wooden table'', (2) ``Look at the multiple small PC monitors on the white desk near the bookshelf'', and (3) ``Check the microwave oven''.}
  \label{figure:advanced-exp}
\end{figure}

\subsection{Advanced Experiment} \label{subsec:advanced-exp}
\switchlanguage%
{%
  We conducted a series of navigation experiments in a larger space.
  By accepting continuous instructions in a wide space, it demonstrates the possibility of more practical navigation.
  The instructions were changed to the following: (1) ``Look at the large TV display on the wooden table'', (2) ``Look at the multiple small PC monitors on the white desk near the bookshelf'', (3) ``Check the microwave oven''.
  In this experiment, we use the proposed method \textbf{ALL}.
  The experimental environment and the trajectory of the robot are shown in \figref{figure:advanced-traj}.
  Here, the initial position ``origin'' of the robot and the center of the target position ``target'' are shown.
  However, as described in \secref{subsec:basic-exp}, the target position has some width, so ``target'' refers to only a rough position.
  The picture seen from ``view-1'' and ``view-2'' shown in \figref{figure:advanced-traj}, and the omnidirectional image of the camera are shown in \figref{figure:advanced-exp}.
  Note that \textcircled{\scriptsize{1}}--\textcircled{\scriptsize{5}} in \figref{figure:advanced-traj} correspond to those in \figref{figure:advanced-exp}.
  Both (1) and (2) deal with the same class of objects such as the display and monitor, but by adding information such as wooden table, bookshelf, white desk, etc., they are correctly distinguished and intended navigation is possible at \textcircled{\scriptsize{2}} and \textcircled{\scriptsize{3}}.
  As for (3), the microwave oven is not visible at \textcircled{\scriptsize{3}}, but the robot is heading toward the kitchen at \textcircled{\scriptsize{4}} where the microwave oven is usually located.
  Then the robot actually finds the microwave in the kitchen and stops in front of it at \textcircled{\scriptsize{5}}.
}%
{%
  より発展的な広い空間での一連のnavigation実験を行った.
  広い空間で連続的な指示を受け付けることで, より実用的なnavigationが可能であることを示す.
  指示を(1) ``Look at the large TV display on the wooden table'', (2) ``Look at the multiple small PC monitors on the white desk near the bookshelf'', (3) ``Check the microwave oven''の順で変化させ, その際の挙動を確認する.
  なお, 本実験では提案手法である\textbf{ALL}を用いている.
  実験環境とロボットの軌跡を\figref{figure:advanced-traj}に示す.
  ここでは, ロボットの初期位置``origin''と指令位置の中心``target''を示しているが, \secref{subsec:basic-exp}と同様に, 指令位置は幅を持っているため, ``target''あくまでも目安である.
  \figref{figure:advanced-traj}に示したview-1, view-2から見た動作, 全天球カメラの視野画像の様子を\figref{figure:advanced-exp}に示す.
  なお, \figref{figure:advanced-traj}の\textcircled{\scriptsize{1}}-\textcircled{\scriptsize{5}}は\figref{figure:advanced-exp}に対応している.
  (1)と(2)は, 両者ともdisplayやmonitorといった同じクラスの物体を扱っているが, そこにwooden tableやbookshelf, white deskなどの付加情報を追加することで, それらを正しく見分けて\figref{figure:advanced-exp}の\textcircled{\scriptsize{2}}と\textcircled{\scriptsize{3}}のようにnavigationすることが可能であった.
  また, (3)については, \textcircled{\scriptsize{3}}の時点ではmicrowave ovenが見えていないが, \textcircled{\scriptsize{4}}のようにそれらが基本的に置いてあるkitchenの方にロボットが向かっている
  その後, kitchenで実際にmicrowaveを見つけ, 最終的に\textcircled{\scriptsize{5}}でmicrowaveの前で停止した.
}%

\section{Discussion} \label{sec:discussion}
\switchlanguage%
{%
  The results obtained are summarized below.
  First, the basic experiment shows that the robot can move to the appropriate position from linguistic instructions by splitting the expanded omnidirectional image, applying multiple pre-trained vision-language models, and controlling the wheels reflexively.
  \textbf{ALL}, which combines CLIP and Detic outputs, enables navigation with higher accuracy compared to either \textbf{CLIP} or \textbf{Detic}.
  As can be seen from the preliminary experiment in \secref{subsec:vlm}, CLIP alone can extract only global features of the image, while Detic alone can extract only local features of the image.
  The combination of these two models enables stable navigation.
  In addition, the robot moved without hesitation in the case of \textbf{CLIP}, while it has shown meandering movements in the case of \textbf{Detic}.
  This may be because the recognition results of Detic differ greatly from step to step.
  Next, the advanced experiment shows that navigation is possible even when the linguistic instructions are changed continuously.
  By adding various modifications to the linguistic instructions, it is possible to distinguish between similar objects and move in the appropriate direction.
  Also, we found that even when the robot is instructed to move toward an object that is not directly visible, it first moves in the direction in which the object is generally expected to exist, finds the indicated object, and then moves toward it.
  This is an interesting feature of the proposed system.

  Limitations and future prospects of this study are described.
  First, several parameters in our method must be manually tuned.
  In particular, the number of segmentations of the omnidirectional image is important, since increasing the number of segmentations increases the inference time while increasing the accuracy of the target direction, which is a trade-off.
  On the other hand, once the desired frequency for control is determined, it is only necessary to set the number of segmentations to fit that frequency, eliminating the need for prior experimentation.
  The degree of overlap of the split images is also important, and appropriate settings are necessary.
  Second, there is a limit to the types of instructions that can be recognized by the vision-language models.
  When there are multiple objects or environments that match the instruction, the model cannot navigate the robot well without appropriate modifications.
  In the case of \secref{subsec:advanced-exp}, we observed that the robot would get lost and constantly rotate unless the instruction was modified by words such as ``wooden table'' or ``bookshelf''.
  There is also a problem that the robot does not recognize negations well.
  The accuracy also varies depending on the size of the target object.
  These problems are currently open questions for general vision-language models, and we believe that more advanced navigation will become possible in the future with the development of these models.
  Third, navigation in more complex spaces is currently difficult for our method.
  There is no problem if the shape of the navigation area is convex, but it may be difficult to perform proper navigation if the shape is concave.
  In such cases, a possible solution is to guide the robot to the desired position by changing the instructions for intermediate landmarks.
  Also, we might be able to solve this problem if we can extract two types of information from language instructions: the direction to head towards and the direction to avoid.
  In the future, we would like to make more advanced reflex-based motion generation possible by adding exploratory behaviors to avoid obstacles and repetitive actions.

  This study is the first example to demonstrate the compatibility of the omnidirectional camera and the vision-language models for reflex-based open-vocabulary navigation by using an actual mobile robot.
  The idea itself is simple, but by using multiple vision-language models in a split omnidirectional camera image, it is possible to determine features of the surroundings both globally and locally, and to perform advanced navigation.
  We believe that this idea has potential applications not only for service robots in daily life, but also for complex and dangerous disaster sites.
}%
{%
  得られた結果についてまとめる.
  まず基礎実験から, 全天球カメラ画像を分割し事前学習済み視覚-言語モデルを適用, 反射的に車輪を制御するのみで, 言語指示に適切な位置にロボットが移動可能なことがわかった.
  この際, CLIPとDeticの出力を合わせる\textbf{ALL}が, CLIPやDetic一方のみを使う場合よりも高い精度でNavigationを可能にする.
  \secref{subsec:vlm}の予備実験からも見て取れる通り, CLIPのみでは画像の全体的な特徴量しか抽出できず, Deticのみでは画像の局所的な特徴量しか抽出できない.
  この2つをかけ合わせて用いることで, 安定したnavigationが可能となっている.
  また, 動きの特徴として, CLIPのみの場合は動きに迷いがなく, Deticのみの場合は蛇行するような動きが多かった.
  これはDeticの認識結果が毎ステップで大きく異なることが原因であったと考えられる.
  次に応用実験から, 指示を連続的に変化させても適切にnavigationが可能であることがわかった.
  特に, 似た物体でも指示に多様な修飾をつけることでそれらを見分け適切な方向に移動することができる.
  また, 直接見えない物体に向かう指示でも, 初めにその物体が一般的に存在するであろう方向に動き, その後指示物体を見つけそちらに向かうという, まず全体を見て徐々に局所的な方向に向かうという高度な動作が生成されることがわかった.

  本研究の限界と今後の展望について述べる.
  まず, 本研究の手法にはいくつかのパラメータが存在する.
  特に全天球画像の分割数は重要で, 分割数を増やすと向かう方向が正確になる一方で推論時間が増えてしまうというトレードオフがある.
  また, 分割画像をどの程度オーバーラップさせるかも重要であり, 適切な設定が必要である.
  次に, 視覚-言語モデルで認識できる指示には限界が存在する.
  指示に合う物体や環境が複数ある場合, 適切な修飾がないとうまくナビゲーションすることができない.
  \secref{subsec:advanced-exp}でも, ``wooden table''や``bookshelf''等の単語により修飾しないとロボットが迷って常に回転するといった動作が見て取れた.
  また, ``not''のような否定をうまく認識してくれないという問題や, 指示物体の大きさによっても精度が変化するといった問題もある.
  これらは一般的な視覚-言語モデルの問題にも通ずるところがあり, 今後モデルの発展と伴により高度なnavigationが可能になると考えている.
  最後に, 本手法は現状, より複雑な空間におけるnavigationは難しい.
  移動可能な空間が凸な形状の場合は問題ないが, 凹な形状の場合, 適切な動作が難しい場合も存在する.
  その場合は, 現状ランドマークごとに指示文を変化させながら, 最終的に到達して欲しい位置まで誘導するという方法が考えられる.
  言語指示から, 向かうべき方向と, 向かうべきでない方向という2種類の情報を取り出すことができれば, この問題を解決することができるかもしれない.
  今後, 障害物を適切に避け, 同じ動作を避けるように探索行動を入れるなどして, より高度な反射型動作生成を可能にしていきたい.

  本研究は, 反射型Open-Vocabulary Navigationに対する全天球カメラと視覚-言語モデルの相性の良さを初めて実機動作から示した研究である.
  そのアイデア自体は非常にシンプルであるが, 全天球カメラ画像を分割し複数の視覚-言語モデルを利用することで周囲の特徴を全体的かつ局所的に判断し, 高度なナビゲーションが可能になる.
  本アイデアは日常生活やレストランを含むサービスロボットだけでなく, 複雑で危険な災害現場等にも応用可能性があると考え, その発展を期待する.
}%

\section{Conclusion} \label{sec:conclusion}
\switchlanguage%
{%
  In this study, we have described a reflex-based open-vocabulary navigation system using an omnidirectional camera and multiple pre-trained vision-language models.
  It does not require any prior knowledge, including map construction and learning.
  By using an omnidirectional camera, the robot can obtain information about its surroundings as a whole, and by splitting the image and applying pre-trained vision-language models to each image, the robot can determine which direction is most consistent with the current linguistic instructions.
  By mapping this direction to the robot's movements, we show that open-vocabulary navigation can be constructed in a simple reflex-based form.
  Moreover, by using multiple types of vision-language models, the robot can respond to a wider variety of linguistic instructions.
  We expect various applications of this idea in the future.
  On the other hand, there are cases where the system does not work well depending on the structure of the room or the objects to be recognized, and we would like to consider a more adaptive robot system that can respond to such situations without the need for prior knowledge.
}%
{%
  本研究では, 全天球画像と事前学習済み視覚-言語モデルを用いたSLAMや学習, 事前知識を必要としないシンプルな反射型Open Vocabulary Navigationについて述べた.
  全天球画像を用いることで周囲全体の情報を取得し, これを分割して事前学習済み視覚-言語モデルを適用することで, どの方向が現在の指示に最も合致しているかを調べることができる.
  それをロボットの動作にマッピングすることで, 言語指示によるナビゲーションが非常にシンプルな反射型の形で構成可能であることを示した.
  また, 複数種類の視覚-言語モデルを用いることでより多様な言語指示に対応でき, 本アイデアの今後の応用が期待される.
  一方で, 部屋の構造や認識物体によっては上手く行かない例もあり, 今後学習や事前知識を必要とせずともそれらに対応可能なよりシンプルで適応的なシステム構成を考えていきたい.
}%

{
  \bibliographystyle{junsrt}
  \bibliography{main}
}

\end{document}